\pdfoutput=1
\documentclass[11pt,a4paper]{article}

\usepackage[utf8]{inputenc}
\usepackage[T1]{fontenc}
\usepackage{newtxtext}
\usepackage{newtxmath}
\newcommand{\eyCoverTitleFont}{\rmfamily}
\newcommand{\eyCoverMetaFont}{\sffamily}
\usepackage[a4paper,margin=1in]{geometry}
\usepackage{graphicx}
\usepackage[protrusion=true,expansion=false]{microtype}
\usepackage{booktabs}
\usepackage{tabularx}
\usepackage{multirow}
\usepackage{array}
\usepackage{enumitem}
\usepackage{xcolor}
\usepackage{tikz}
\usetikzlibrary{arrows.meta,positioning,shapes.geometric,fit,calc,decorations.markings,patterns,backgrounds}
\usepackage[authoryear,round]{natbib}
\usepackage{hyperref}
\usepackage{cleveref}
\usepackage{float}
\usepackage{fancyhdr}

\definecolor{eyblue}{RGB}{38,100,168}
\definecolor{eyteal}{RGB}{0,140,130}
\definecolor{eyorange}{RGB}{220,130,20}
\definecolor{eypurple}{RGB}{120,70,160}
\definecolor{eyred}{RGB}{195,55,55}
\definecolor{eygray}{RGB}{240,243,248}
\definecolor{eylightblue}{RGB}{220,235,252}
\definecolor{eylightteal}{RGB}{218,245,242}
\definecolor{eyforestgreen}{HTML}{1D3B2A}
\definecolor{eysectiongreen}{HTML}{1D3B2A}
\definecolor{eytitletext}{HTML}{202124}
\definecolor{eytitleaccent}{HTML}{A7B8AE}
\definecolor{eytitlebox}{HTML}{F7F8F5}
\definecolor{eylightorange}{RGB}{255,240,218}
\definecolor{eylightpurple}{RGB}{238,228,248}
\definecolor{eylightred}{RGB}{252,228,228}
\definecolor{darkgray}{RGB}{80,80,80}

\hypersetup{
  colorlinks=true,
  hypertexnames=false,
  linkcolor=eysectiongreen,
  citecolor=eysectiongreen,
  urlcolor=eyforestgreen,
  pdftitle={Eywa: Provenance-Grounded Long-Term Memory for AI Agents},
  pdfauthor={Resham Joshi},
  pdfsubject={Long-term AI agent memory},
  pdfkeywords={agent memory, long-term memory, retrieval, provenance, LoCoMo, LongMemEval, BEAM}
}
\providecommand{\doi}[1]{\href{https://doi.org/#1}{doi: \nolinkurl{#1}}}

\renewcommand{\headrulewidth}{0.4pt}

\renewcommand{\headrule}{\hbox to\headwidth{\color{eytitleaccent}\leaders\hrule height \headrulewidth\hfill}}

\tikzset{
  roundbox/.style={draw=#1, rounded corners=4pt, fill=#1!8, minimum height=10mm,
    text width=28mm, align=center, font=\small, line width=0.7pt},
  roundbox/.default=eyblue,
  smallbox/.style={draw=#1, rounded corners=3pt, fill=#1!8, minimum height=8mm,
    text width=22mm, align=center, font=\footnotesize, line width=0.7pt},
  smallbox/.default=eyblue,
  wideblock/.style={draw=#1, rounded corners=4pt, fill=#1!8, minimum height=10mm,
    text width=38mm, align=center, font=\small, line width=0.7pt},
  wideblock/.default=eyblue,
  decision/.style={draw=eyorange, diamond, aspect=2.2, align=center,
    fill=eyorange!8, inner sep=1pt, text width=22mm, font=\footnotesize, line width=0.7pt},
  arr/.style={-{Latex[length=2.5mm,width=1.8mm]}, line width=0.8pt, #1},
  arr/.default=darkgray,
  dasharr/.style={-{Latex[length=2.5mm,width=1.8mm]}, line width=0.7pt, dashed, #1},
  dasharr/.default=darkgray!60,
  groupbox/.style={draw=#1!50, rounded corners=6pt, fill=#1!4,
    inner sep=5pt, line width=0.6pt, dashed},
  groupbox/.default=eyblue,
  label/.style={font=\scriptsize\sffamily, text=darkgray},
}

\newcolumntype{R}{>{\raggedleft\arraybackslash}p{12mm}}
\newcolumntype{C}{>{\centering\arraybackslash}p{12mm}}
\newcommand{\eyFigureLogo}{\par\vspace{-5pt}\hfill\includegraphics[width=0.095\linewidth]{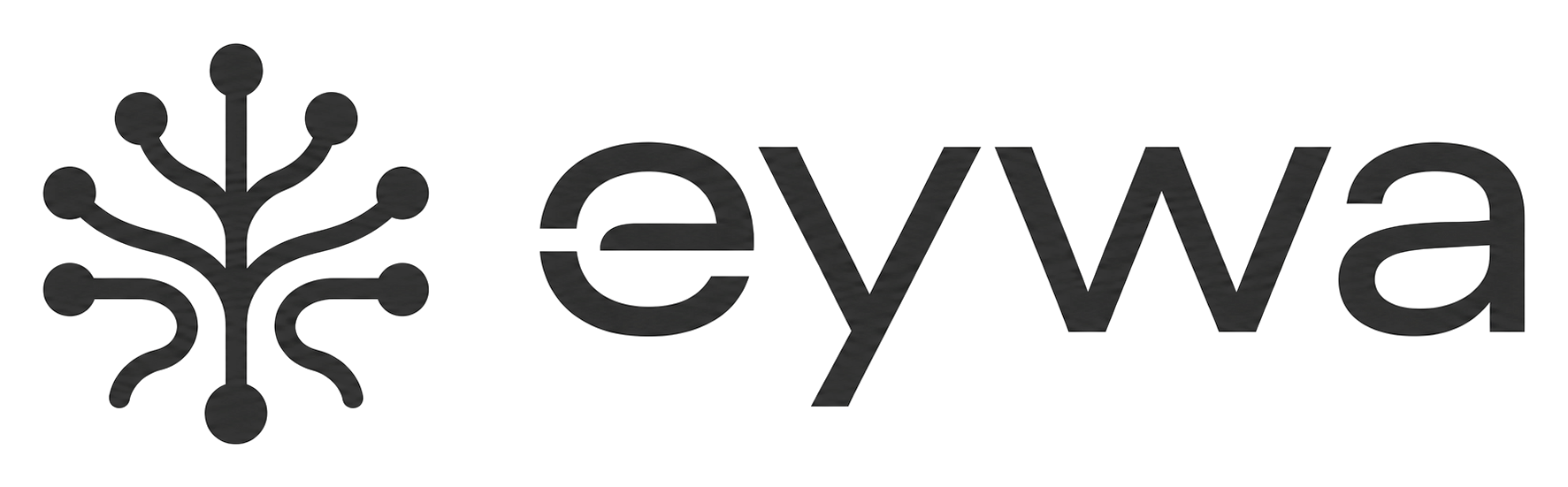}\par\vspace{-3pt}}

\makeatletter
\renewcommand\section{\@startsection{section}{1}{\z@}%
  {-3.5ex \@plus -1ex \@minus -.2ex}%
  {2.3ex \@plus .2ex}%
  {\normalfont\Large\bfseries\color{eysectiongreen}}}
\renewcommand\subsection{\@startsection{subsection}{2}{\z@}%
  {-3.25ex\@plus -1ex \@minus -.2ex}%
  {1.5ex \@plus .2ex}%
  {\normalfont\large\bfseries\color{eysectiongreen}}}
\renewcommand\paragraph{\@startsection{paragraph}{4}{\z@}%
  {3.25ex \@plus1ex \@minus.2ex}%
  {-1em}%
  {\normalfont\normalsize\bfseries\color{eysectiongreen}}}
\renewcommand{\@seccntformat}[1]{\csname the#1\endcsname.\enspace}
\makeatother

\title{\textbf{Eywa: Provenance-Grounded Long-Term Memory for AI Agents}}

\author{
  Resham Joshi\\
  \texttt{research@eywa.to}\\
  \url{https://eywa.to}
}

\date{May 2026}

\begin{document}
\thispagestyle{empty}

\begin{center}
\vspace*{-0.35cm}
\begin{tikzpicture}
\node[
  fill=eytitlebox,
  draw=eytitleaccent!40,
  rounded corners=7pt,
  inner xsep=9pt,
  inner ysep=12pt,
  text width=0.94\textwidth,
  align=left
]{
  \begin{center}
    {\eyCoverTitleFont\fontsize{20}{24}\selectfont\bfseries\color{eyforestgreen}
    Eywa: Provenance-Grounded Long-Term Memory for AI Agents\par}
    \vspace{7pt}
    {\eyCoverMetaFont\normalsize\bfseries Resham Joshi\par}
	    \vspace{3pt}
		    {\eyCoverMetaFont\small research@eywa.to\par}
		    \vspace{2pt}
		    {\eyCoverMetaFont\small\href{https://eywa.to}{https://eywa.to}\par}
		    \vspace{5pt}
		    \includegraphics[width=0.22\textwidth]{figures/eywa_logo_wordmark.png}\par
		    \vspace{4pt}
		    {\eyCoverMetaFont\small Benchmarks and artifacts: \href{https://eywa.to/research}{https://eywa.to/research}\par}
		  \end{center}

	  \vspace{9pt}
  {\small
  \color{eytitletext}
  \noindent AI agents that persist across sessions need memory they can retrieve, audit, update, and erase. Existing memory systems often collapse source evidence, extracted facts, retrieved context, and answer policy into one opaque prompt path, making failures difficult to diagnose: a wrong answer may come from missing evidence, unsupported extraction, stale state, retrieval loss, or answer-model behavior. We present \textbf{Eywa}, a provenance-grounded memory architecture built around \emph{evidence before belief}. Eywa stores immutable source evidence before deriving canonical facts, validates extracted memories against typed signals and source support, and retrieves bounded memory context through a deterministic multi-route read path with zero LLM calls inside retrieval. Retrieved context is returned separately from answer instructions, allowing the same memory substrate to be evaluated across frontier, budget, and local answer models. Under a frozen, artifact-recorded retrieval configuration, Eywa reaches \textbf{90.19\%} judge accuracy on the LoCoMo C1--C4 split with Claude Sonnet 4.6 write and QA roles. On LongMemEval-S, it reaches \textbf{88.2\%} retrieval-sufficiency accuracy. On BEAM, a 700-question technical-memory stress benchmark, it reaches \textbf{81.45\%} mean nugget score and \textbf{85.29\%} pass@score$\geq$0.5. Full per-question artifacts, including questions, gold answers, model answers, retrieved context, and labels, are published at \url{https://eywa.to/research}.
  }
};
\end{tikzpicture}
\end{center}

\section{Introduction}

Human memory is not merely an archive of facts. It supports self-continuity, social coordination, and future-oriented action: autobiographical memory helps people maintain identity, learn from prior experience, make decisions, and sustain relationships over time~\citep{conway2000selfmemory,bluck2005tale,schacter2007prospective}. Conversation also depends on memory. Speakers build common ground by remembering what has already been established with a particular partner, what remains uncertain, and what can safely be assumed in later exchanges~\citep{clark1991grounding}. For AI agents that interact with users across days, projects, and sessions, the analogous requirement is not biological fidelity, but functional continuity: an agent must preserve important past evidence, update stale beliefs, retrieve relevant context, and explain why a remembered claim is justified.

Large language models have made this requirement more visible. Modern LLMs can produce fluent and contextually appropriate responses within a prompt, but their native computation is bounded by a finite working context. Scaling this window is valuable, and frontier systems such as Gemini 1.5 demonstrate impressive long-context retrieval over very large inputs~\citep{gemini15}. Yet a larger window is not the same thing as a memory architecture. Transformer attention remains costly at long sequence lengths~\citep{flashattention}, and empirical studies show that long-context models can fail to use information reliably when relevant evidence appears in the middle of the prompt or when tasks exceed the model's effective context size~\citep{lostmiddle,ruler}. More importantly, context extension alone does not decide what should be stored, what should be revised, what should be forgotten, or how a user can audit and erase remembered information.

Prior work has therefore moved from stateless prompting toward explicit agent memory. Generative Agents introduced memory streams, reflection, and planning as core mechanisms for believable long-horizon behavior~\citep{generativeagents}. Reflexion stores verbal feedback to improve later agent attempts~\citep{reflexion}. MemoryBank maintains long-term user memory for conversational personalization~\citep{memorybank}, while MemGPT frames memory management as a virtual-context problem in which information is moved between working and archival stores~\citep{memgpt}. Recent benchmarks such as LoCoMo and LongMemEval make this problem measurable across multi-session dialogue, temporal reasoning, knowledge updates, abstention, and cross-session retrieval~\citep{locomo,longmemeval}. Production-oriented and research systems including Mem0, Supermemory, Zep/Graphiti, A-MEM, MemoryOS, and graph-based RAG methods further show that external memory, temporal structure, and graph organization can improve long-horizon agents~\citep{mem0paper,supermemory,zep,amem,memoryos,hipporag}.

Despite this progress, a practical failure remains under-specified: when a memory-backed answer is wrong, it is often unclear which layer failed. The original conversation may have contained the right evidence but the extractor may have omitted it. The extractor may have created a plausible but unsupported fact. Retrieval may have found the fact but ranked it below the context budget. The answer model may have received the evidence and still refused, overgeneralized, or answered for the wrong person. End-to-end scores collapse these cases into a single number, making it difficult to improve a system without guessing.

This paper presents \textbf{Eywa}, a provenance-grounded long-term memory architecture for AI agents. Eywa is built around the principle of \emph{evidence before belief}: raw conversational evidence is preserved before any LLM-derived fact becomes canonical memory. The write path stores immutable evidence, detects typed signals such as dates, entities, decisions, corrections, and hard anchors, and validates extracted facts against their source evidence before they are promoted. Canonical facts remain linked to the evidence from which they were derived, allowing later inspection, repair, and deletion.

Provenance establishes source support, not external truth. If a user says something false, ambiguous, or sarcastic, Eywa can preserve where the claim came from and whether the extracted belief is supported by that source, but world-level truth verification remains outside the memory layer.

Eywa also separates memory retrieval from answer synthesis. Its read path performs deterministic multi-route retrieval over canonical facts, observations, temporal metadata, entity scope, keyword search, and vector similarity without LLM calls on the retrieval path. The retrieved evidence is then passed to an answer model under an explicit answer policy. This separation makes it possible to measure the memory system independently from the model that verbalizes the final response: the same retrieval architecture can serve a frontier API model, a small local model, or a stricter refusal policy.

We evaluate Eywa on LoCoMo C1--C4, LongMemEval-S, and BEAM using role-explicit, trace-preserving runs. LongMemEval-S and BEAM test different memory behaviors, so we report them under their own protocols rather than reducing them to a single average. Across these settings, the central result is not only that Eywa scores well, but that each score can be inspected question by question through the evidence, facts, retrieved context, answer, and judge label that produced it.

\newpage
\paragraph{Contributions.}
\begin{enumerate}[leftmargin=*,topsep=3pt,itemsep=1pt]
  \item A provenance-grounded memory model that separates immutable source evidence from canonical LLM-derived beliefs.
  \item A two-tier write path with typed signal detection and hard-anchor validation for reducing unsupported extracted memories.
  \item A deterministic multi-route read path that retrieves memory context without LLM calls and preserves traceability to source evidence.
  \item A model-separable answer interface showing that the same retrieved evidence can be consumed by hosted frontier, hosted budget, and local models.
  \item A trace-level evaluation protocol that reports write-path, QA, and judge roles separately and preserves per-question provenance for independent inspection.
\end{enumerate}

\section{Background and Related Work}

A recent survey~\citep{memorysurvey} organizes agent memory along three axes: \emph{forms} (token-level, parametric, latent), \emph{functions} (factual, experiential, working), and \emph{dynamics} (formation, evolution, retrieval). This taxonomy is useful because it separates the representation of memory from the operations that maintain it. Eywa follows this distinction: it treats memory not as a single store, but as a lifecycle that includes evidence capture, fact formation, revision, retrieval, and answer-time use.

\paragraph{Long context and retrieval limits.}
One response to long-horizon interaction is to increase the model's context window. Gemini 1.5 demonstrates that million-token contexts can substantially expand what a model can condition on~\citep{gemini15}. At the same time, long context does not remove the need for memory management. Attention remains computationally expensive at long sequence lengths~\citep{flashattention}, and studies such as Lost in the Middle and RULER show that models can still fail to retrieve or use relevant evidence reliably inside long prompts~\citep{lostmiddle,ruler}. These findings motivate memory systems that select, structure, and retrieve evidence rather than replaying all prior interaction.

\paragraph{Experience and reflection memory.}
Generative Agents introduced a memory stream in which observations are retrieved, reflected on, and used for planning~\citep{generativeagents}. Reflexion stores verbal feedback from failed attempts so that later agent behavior can improve without parameter updates~\citep{reflexion}. MemoryBank extends this line to long-term conversational personalization, including mechanisms for updating and forgetting user memories over time~\citep{memorybank}. These systems establish that memory is not just storage; it is an active control surface for behavior over repeated interactions.

\paragraph{Hierarchical and operating-system views.}
MemGPT frames memory management as a virtual-context problem, using an operating-system analogy in which information moves between working context and archival storage~\citep{memgpt}. MemoryOS develops a related hierarchy with short-term, mid-term, and long-term personal memory, together with update and retrieval modules inspired by storage management~\citep{memoryos}. These approaches emphasize capacity management: deciding what should remain immediately available, what should be compressed or paged out, and what should be retrieved later.

\paragraph{Production memory layers.}
Mem0 presents a practical memory layer for extracting, updating, and retrieving salient conversational memories, with a graph-enhanced variant for relational structure~\citep{mem0paper}. Supermemory exposes an open-source memory engine and hosted API for adding, searching, and injecting persistent user context across AI tools, with SDKs, MCP support, and application connectors~\citep{supermemory}. LangMem provides memory management components for LangGraph-based agents~\citep{langmem}. OpenAI's ChatGPT memory exposes saved-memory and chat-history reference controls as a product feature, while its internal memory architecture is not publicly specified~\citep{openaimem}. This line of work is important because it treats memory as infrastructure that must operate under latency, cost, privacy, and product-control constraints.

\paragraph{Graph and temporally structured retrieval.}
Graph-structured memory is a natural fit for multi-hop and temporally evolving information. Zep/Graphiti builds a temporally aware knowledge graph with a bi-temporal model tracking both real-world validity and system ingestion time~\citep{zep}. GraphRAG-style methods show how graph structure can support query-focused retrieval and summarization over large knowledge sources~\citep{graphrag}. HippoRAG brings a complementary perspective from neurobiologically inspired retrieval, using graph structure and hippocampal-indexing ideas for long-term memory over knowledge sources~\citep{hipporag}. A-MEM uses a Zettelkasten-inspired process in which an LLM creates, links, and evolves memory notes over time~\citep{amem}. Together, these systems show that relation structure, temporal validity, and memory evolution are central to long-term agent behavior.

\paragraph{Recent agent-memory architectures.}
Hindsight organizes agent memory into separate logical networks for world facts, experiences, observations, and opinions, with explicit retain, recall, and reflect operations~\citep{hindsight}. This emphasizes epistemic role separation and profile-conditioned reasoning. True Memory argues that extraction at ingestion is the wrong primitive and instead centers retrieval over verbatim preserved events~\citep{truememory}. Eywa is closest to these systems in treating memory as an architecture rather than a vector-store feature. Its distinction is provenance-grounded evidence separation: source evidence is stored as the authoritative substrate, while extracted facts are validated, revisable indexes over that evidence rather than a lossy replacement for it. Eywa also keeps retrieval LLM-free and returns answer policy separately from retrieved memory context.

\paragraph{Benchmarks and evaluation.}
LoCoMo~\citep{locomo} evaluates very long-term conversational memory over long, multi-session dialogues. Its public benchmark release includes 10 conversations with 1{,}986 QA questions: 841 single-hop, 282 multi-hop, 321 temporal-reasoning, 96 open-domain-knowledge, and 446 adversarial questions. Following the published-result convention used by recent memory-system result tables, we report the four non-adversarial categories as the main C1--C4 comparison split. LongMemEval~\citep{longmemeval} tests five core long-term memory abilities: information extraction, multi-session reasoning, knowledge updates, temporal reasoning, and abstention. These benchmarks are valuable because they expose memory as an end-to-end behavior, but their final-answer scores do not by themselves identify whether a failure came from extraction, retrieval, context packing, answer policy, or judging.

\paragraph{Infrastructure and governance.}
Memory quality is only one part of a deployable system. Pancake~\citep{pancake} addresses vector-search throughput for multi-agent workloads through hierarchical caching and GPU coordination. Concurrent work on memory governance~\citep{animesis} studies constitutional frameworks for long-lived digital citizens. These directions are complementary to semantic memory architectures because a useful memory system must be accurate, scalable, inspectable, and controllable.

\paragraph{Design positioning.}
Eywa is positioned within this memory-management layer. It makes three architectural choices. First, it separates immutable source evidence from canonical LLM-derived beliefs so that memories can be audited and repaired. Second, it keeps retrieval deterministic and LLM-free, using multiple retrieval routes, temporal metadata, entity scope, and provenance-preserving context packing. Third, it separates memory delivery from answer policy, allowing the same memory architecture to support local, low-cost, or frontier answer models. The goal is not to replace prior memory systems wholesale, but to make the write, read, and answer stages easier to inspect and improve independently.

\section{Problem Definition and Design Goals}
\label{sec:problem}

We define long-term agent memory as an external system that transforms an ongoing stream of interaction into retrievable, auditable context for future answers. Let a user's interaction history be a sequence of turns
\[
H_u = \{x_1,\ldots,x_n\}, \quad x_i = (u, r_i, t_i, s_i),
\]
where \(u\) is the user or tenant scope, \(r_i\) is the speaker role, \(t_i\) is the timestamp or session time, and \(s_i\) is the raw utterance or event text. A memory system receives \(H_u\), stores a set of source evidence records \(E_u\), derives a set of structured memories or beliefs \(B_u\), and, for a later query \(q\), returns a bounded context \(C(q,u,t)\) that an answer model can use to produce either an answer \(a\) or an abstention.

This formulation differs from ordinary document retrieval in four ways. First, the corpus is not static: users correct themselves, change preferences, and introduce facts whose validity depends on time. Second, the relevant unit is often not a whole document but a small conversational event, preference, decision, correction, or relation. Third, answerability depends on user scope and provenance: a fact about one person should not answer a question about another, and an extracted claim should remain traceable to its source. Fourth, benchmark and product quality depend not only on whether a relevant item can be found, but also on whether the answer model uses it under the chosen policy. LoCoMo and LongMemEval make these pressures explicit through multi-session reasoning, temporal questions, knowledge updates, abstention, and long-range retrieval tasks~\citep{locomo,longmemeval}. Long-context evaluations further show why simply appending more history is insufficient: models can fail to use evidence reliably as prompts grow or as relevant evidence moves away from favorable positions~\citep{lostmiddle,ruler}.

\paragraph{Memory objects.}
Eywa separates three objects that are often collapsed in memory systems.

\begin{enumerate}[leftmargin=*,topsep=3pt,itemsep=1pt]
  \item \textbf{Evidence} is immutable source material: the original user turn, assistant proposal, observation, timestamp, role, and metadata. Evidence may be sensitive and must remain deletable by user scope.
  \item \textbf{Signals} are typed detections over evidence, such as people, organizations, places, dates, versions, URLs, decisions, corrections, approvals, rejections, or other hard anchors.
  \item \textbf{Beliefs} are LLM-derived memories that may be canonical profile facts, episodic observations, temporal events, or relations. A belief is valid only to the extent that it is supported by one or more evidence records.
\end{enumerate}

The central invariant is that evidence precedes belief. A system may revise or delete a canonical belief, but it should not silently mutate the raw source from which that belief was derived. This invariant supports auditing, repair, user erasure, and reproducible failure analysis.

\paragraph{Failure taxonomy.}
For a memory-backed answer, an error can originate at several layers. Treating all of them as a single final-answer failure hides the actual engineering problem. We therefore use the following taxonomy throughout the paper. The first seven rows describe system-level failures; the final row describes an evaluation-layer failure in which the metric, rather than the memory system, is the source of disagreement.

\begin{table}[H]
\centering
\small
\setlength{\tabcolsep}{5pt}
\renewcommand{\arraystretch}{1.12}
\begin{tabularx}{\linewidth}{@{}p{0.20\linewidth}X X@{}}
\toprule
\textbf{Failure mode} & \textbf{Definition} & \textbf{Observable symptom} \\
\midrule
Coverage gap & The source conversation contained the information, but the memory store does not contain supporting evidence or a derived belief. & No route can retrieve the answer because the store lacks the needed support. \\
Grounding gap & A derived belief is not supported by its source evidence, or a hard value such as a date, version, amount, URL, or name was changed. & The memory store contains a plausible but source-inconsistent claim. \\
Revision gap & A newer correction, preference change, or superseding event is not applied when retrieving current state. & The system returns both outdated and current facts, or chooses the outdated one. \\
Scope gap & A memory from the wrong user, speaker, person, or entity is used. & The answer is semantically related but belongs to someone else. \\
Temporal gap & The memory is relevant but selected under the wrong time constraint, ordering, or validity window. & The answer ignores before/after, date range, recency, or historical-state wording. \\
Retrieval gap & The needed belief exists but is not retrieved, is ranked too low, or is removed by context packing. & The database has the answer but the answer model never sees it. \\
Synthesis gap & The answer model receives sufficient evidence but refuses, overgeneralizes, omits list items, or produces the wrong natural-language answer. & Trace shows supporting context present, but final answer is wrong. \\
Measurement gap & The system answer is semantically supported, but the scoring metric does not credit it, or credits an unhelpful refusal. & Token F1 and LLM judge disagree. \\
\bottomrule
\end{tabularx}
\caption{Failure and evaluation modes for long-term agent memory. The taxonomy is used to separate extraction, retrieval, answer-policy, and metric errors.}
\label{tab:failure-taxonomy}
\end{table}

This taxonomy is intentionally operational. It mirrors the indexing, retrieval, and reading decomposition used in LongMemEval~\citep{longmemeval}, but adds provenance, revision, scope, temporal validity, and measurement distinctions that are important for production memory systems.

\paragraph{Design goals.}
A production memory system should satisfy the following goals. Some goals are related, but we keep them separate because they fail independently in practice: a system can preserve evidence but alter a hard value, retrieve relevant facts but violate entity scope, or deliver the right evidence to an answer model that still refuses.

\begin{enumerate}[leftmargin=*,topsep=3pt,itemsep=1pt]
  \item \textbf{Faithful write path.} The system should capture conversational evidence before deriving beliefs, and derived beliefs should remain linked to source evidence.
  \item \textbf{Grounded extraction.} The system should distinguish soft semantic claims from hard anchors whose values must match the source text exactly.
  \item \textbf{Revision awareness.} The system should represent updates, corrections, rejections, approvals, and superseding facts rather than treating all memories as flat, timeless assertions.
  \item \textbf{Scoped retrieval.} Retrieval should respect user, speaker, person, role, and time constraints, because many memory failures are wrong-scope failures rather than missing-knowledge failures.
  \item \textbf{Bounded context delivery.} The read path should select and order a small amount of high-value context under a token budget instead of replaying the full history.
  \item \textbf{Model separability.} Memory delivery should be separable from answer synthesis so that a local model, hosted model, strict policy, or inference-permissive policy can use the same retrieved context.
  \item \textbf{Auditability and erasure.} Every answerable memory should be traceable to evidence, and every evidence record should remain removable by user scope.
  \item \textbf{Diagnostic evaluation.} Evaluation should report not only final answer quality, but also whether the failure came from coverage, grounding, retrieval, synthesis, or metric behavior.
\end{enumerate}

\paragraph{Non-goals.}
Eywa does not attempt to make the answer model omniscient, to infer unsupported facts, or to store every token forever. It also does not claim that external memory replaces long-context models. Instead, Eywa treats long-context capability as one possible consumer of a better memory substrate. The memory layer is responsible for preserving, validating, retrieving, and explaining evidence; the answer model is responsible for verbalizing that evidence under a chosen policy.

\clearpage
\section{System Architecture}

Eywa is built around three separations: evidence from belief, retrieval from answering, and policy from memory. The memory system does not ask an answer model to decide what should be remembered at query time. It first captures source evidence, derives validated beliefs from that evidence, retrieves a bounded set of support, and returns both memory context and separate answer instructions to the downstream model. This turns long-term memory from opaque prompt state into an auditable system boundary.

\begin{figure}[H]
\centering
\includegraphics[width=\linewidth]{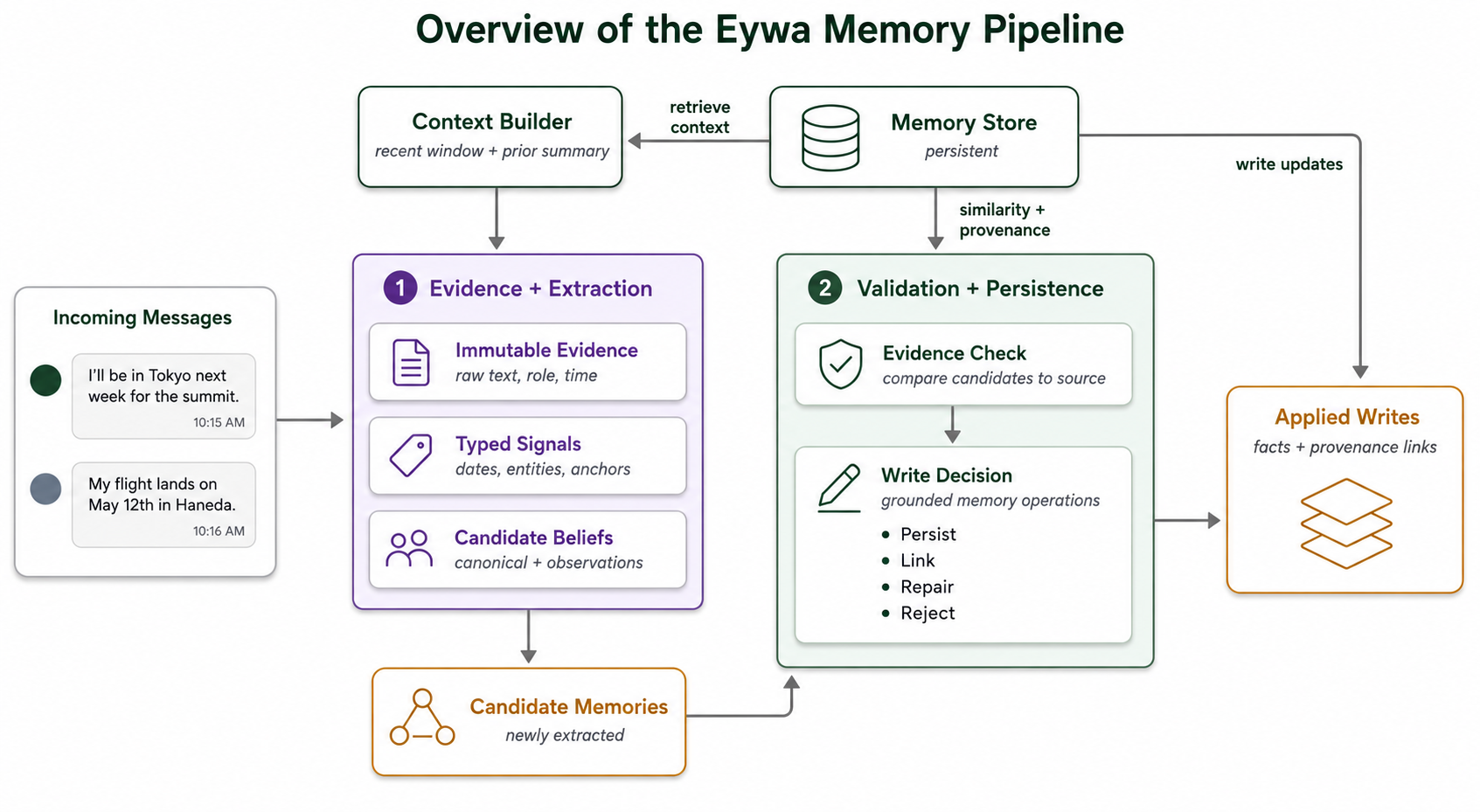}
\eyFigureLogo
\caption{Overview of the Eywa memory pipeline. The write path captures immutable evidence and typed signals before validated memories are persisted. The read path later retrieves bounded, provenance-grounded context for the answer model.}
\label{fig:pipeline-overview}
\end{figure}

The loop begins when user messages routed into memory enter Tier~0 as immutable evidence and typed signals. Tier~1 proposes memories, validates them against the captured source, and writes accepted facts with source metadata or evidence links. At read time, deterministic retrieval routes select a bounded support set from the memory store. Because reads and writes share the same provenance substrate, an answer can be traced back to the evidence that produced the retrieved memory.

\subsection{Write Path: Evidence Before Belief}

The write path operates in two tiers:

\paragraph{Tier 0: Immutable capture.} Every user turn routed to the evidence pipeline is stored as an immutable evidence row. A signal detector extracts typed anchors (dates, entities, monetary values, version numbers, identifiers, URLs, IP addresses, percentages, and quoted strings) without any LLM call. These signals become the deterministic validation substrate against which extracted facts are checked.

\paragraph{Tier 1: Validated extraction.} An LLM extracts candidate facts $\hat{F} = \{\hat{f}_1, \dots, \hat{f}_n\}$ from the evidence window. Each candidate is aligned against Tier~0 evidence through a validation function $V$:
\begin{equation}
  V(\hat{f}, \mathcal{E}, \mathcal{S}) =
  V_{\text{support}}(\hat{f}, \mathcal{E})
  \wedge V_{\text{hard}}(\hat{f}, \mathcal{S})
  \wedge V_{\text{subject}}(\hat{f}, \mathcal{E})
  \wedge V_{\text{act}}(\hat{f}, \mathcal{S})
  \label{eq:validation}
\end{equation}
where $\mathcal{E}$ is the aligned evidence window and $\mathcal{S}$ is the Tier~0 signal set. $V_{\text{support}}$ requires lexical and fuzzy overlap with supporting source text, $V_{\text{hard}}$ verifies deterministic values such as dates, monetary values, quoted strings, identifiers, URLs, IP addresses, and percentages, $V_{\text{subject}}$ checks that the claimed subject appears in the support window or known entities, and $V_{\text{act}}$ preserves negation and uncertainty from memory-act signals. Thus hard anchors are a guardrail, not the whole validator. In a 143-sample two-tier audit, 67.4\% of candidate facts contained no hard anchor; these candidates were still checked by support overlap, subject, and polarity rules. The same audit rejected 11 of 132 candidate facts, most commonly for insufficient source overlap or invented hard values. Validated facts are eligible for canonical write; failed candidates are rejected or marked for repair. In the two-tier write path, each accepted canonical fact $f$ retains a provenance link $\ell(f) \to e$ to its source evidence row $e$.

\subsection{Extraction Is an Index, Not the Memory}

Eywa does not assume that LLM extraction can anticipate every future query. The authoritative memory substrate is the immutable evidence store. Extracted canonical facts are compact, queryable, and revisable beliefs linked back to that evidence; they make retrieval efficient, but they do not replace the source. If a fact is missing, stale, or wrong, the source evidence remains inspectable and can support repair, re-extraction, deletion, or a deeper provenance-rescue path.

This distinction is central to the architecture. Extraction-only systems risk making the extracted item the only durable memory object: if a detail is discarded before the query is known, retrieval cannot recover it later. Eywa instead treats extraction as an auditable index over preserved evidence. The default read path retrieves compact facts for speed, while audit and benchmark traces can expose the source evidence behind those facts. Thus the system can be evaluated both as a practical retrieval layer and as a verifiable memory substrate.

\subsection{Memory Representation and Provenance}

Eywa represents memory as a provenance graph rather than a flat fact table: evidence records are immutable roots, signals are typed annotations over evidence, canonical facts and observations are derived beliefs, and provenance metadata connects beliefs back to the source material that supports them. The implementation realizes this graph over SQLite tables, LanceDB vector indexes, SQLite full-text keyword indexes, and graph relations, but the invariant is representation-independent: raw evidence is never rewritten, while candidate or derived beliefs can be accepted, repaired, superseded, or rejected as new evidence arrives.

In the local-first implementation reported here, SQLite is the authoritative store for evidence, facts, lifecycle status, and provenance metadata. The vector index and graph checkpoint are idempotent projections used for retrieval speed and relational traversal. They are not treated as independent sources of truth: if a projection drifts, the expected repair path is projection rebuild or startup reconciliation from SQLite. A transactional outbox or background reconciler is the appropriate extension for multi-process deployments, but it is not required for the deliberate single-process local-first configuration evaluated in this paper.

\begin{table}[H]
\centering
{\small
\begin{tabular}{@{}l l l p{0.38\linewidth}@{}}
\toprule
\textbf{Object} & \textbf{Mutability} & \textbf{Producer} & \textbf{Role} \\
\midrule
Evidence & Immutable & Conversation ingest & Raw text with role, user, time, and session metadata. \\
Signals & Append-only & Deterministic detectors & Dates, entities, anchors, and memory acts extracted from evidence. \\
Candidates & Transient & Extractor & Proposed canonical or observation memories before validation. \\
Beliefs & Revisable & Validator / writer & Durable facts used by retrieval, with lifecycle and scope metadata. \\
Links & Append-only & Persistence layer & Provenance edges or source metadata from beliefs to evidence and supporting signals. \\
\bottomrule
\end{tabular}}
\caption{Eywa memory object model. Evidence is immutable; beliefs are derived and can carry lifecycle state.}
\label{tab:memory-model}
\end{table}

\subsection{Extraction Modes}

Eywa exposes extraction mode as an architectural choice because different memory tasks require different recall and precision tradeoffs:
\begin{itemize}[leftmargin=*,topsep=2pt,itemsep=1pt]
  \item \textbf{Canonical:} Durable profile facts (``Alice prefers dark mode'') with entity typing and lifecycle tracking.
  \item \textbf{Observation:} High-recall episodic facts (``Alice mentioned going to a pride parade on July 3, 2023'') that preserve event-level detail.
  \item \textbf{Hybrid:} Both paths run for the same user turn. This is the recommended high-recall deployment mode.
  \item \textbf{Structured technical extraction:} Domain-neutral technical memories such as implementation details, library versions, schema details, configuration choices, negative claims, milestones, and metric summaries. These extracted facts are typed separately from profile and episodic memories so the read path can retrieve them without treating code, system design, or benchmark-analysis facts as ordinary user preferences.
\end{itemize}

Canonical extraction supports stable preferences, identity, and current-state memory. Observation extraction supports episodic recall, temporal questions, and open-domain questions where the answer depends on a specific event rather than a durable profile fact. Structured technical extraction preserves implementation-level detail that would otherwise be lost in a conversational summary, including exact tool names, versions, configuration parameters, dates, counts, and explicit ``not implemented'' or ``not using'' claims. Hybrid extraction lets the read path choose between compact canonical state, detailed episodic evidence, and typed technical memories without forcing a single memory representation to serve every question type.

\subsection{Read Path: Multi-Route Deterministic Retrieval}

\begin{figure}[H]
\centering
\includegraphics[width=\linewidth]{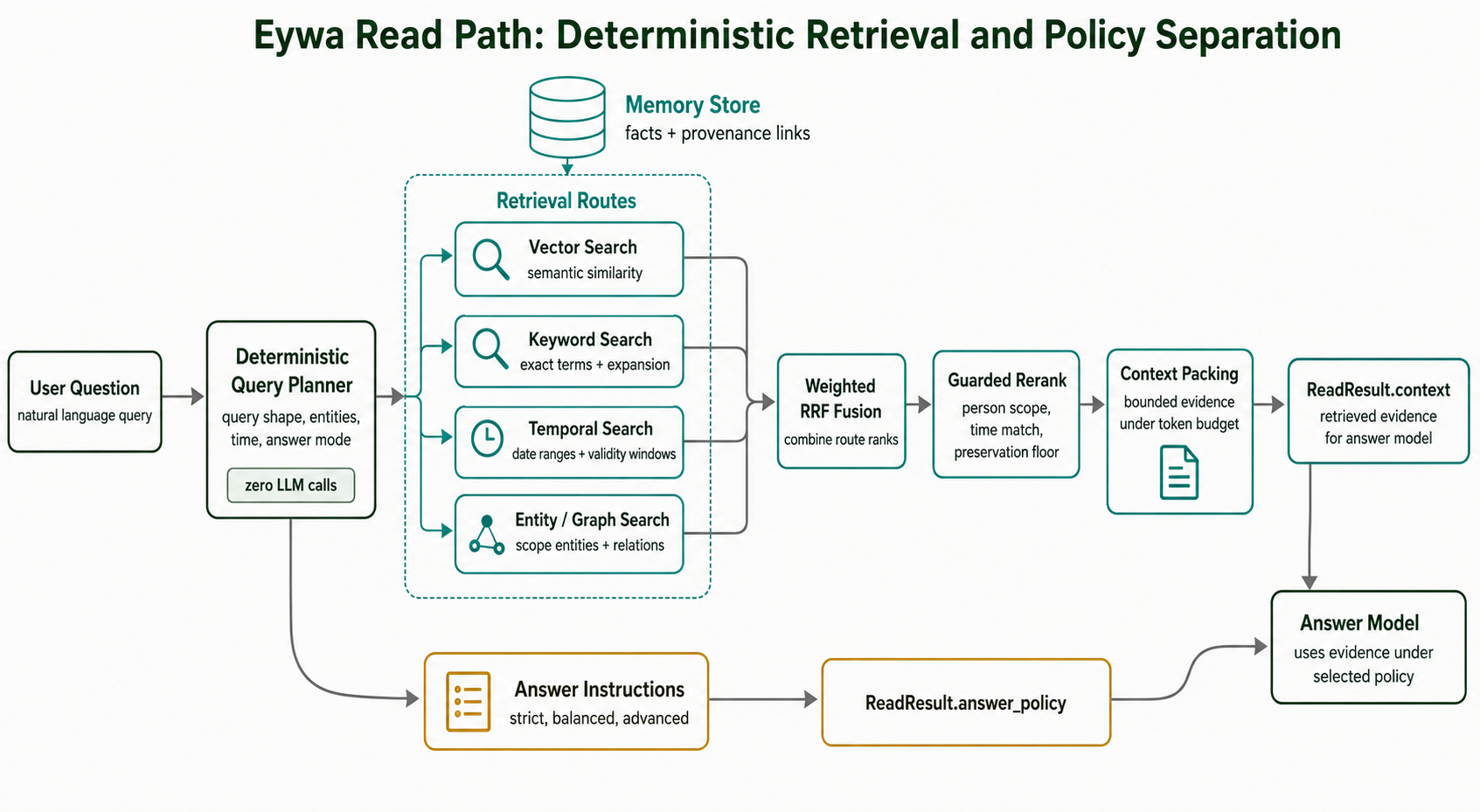}
\eyFigureLogo
\caption{Eywa read path. The query planner is deterministic (no LLM call). Retrieval and support channels contribute candidates with query-shape rule weights. Answer instructions flow from the planner directly, never mixed with memory context.}
\label{fig:read-path}
\end{figure}

The read path (\cref{fig:read-path}) uses zero LLM calls inside memory retrieval. In this paper, ``zero-LLM read'' means no LLM call occurs between receiving a query and returning retrieved memory context: planning, retrieval, fusion, and context packing are deterministic. It does not mean the whole answer pipeline is LLM-free, because the downstream QA model may still be an LLM, and it does not mean memories were written without LLMs. A deterministic query planner analyzes the question's shape (exact, recommendation, inference, count, interval, boolean, list, or factoid), together with temporal and entity cues. It then assigns query-shape rule weights to the active retrieval and support channels:

\begin{itemize}[leftmargin=*,topsep=2pt,itemsep=1pt]
  \item \textbf{Vector ($R_v$):} Embedding cosine similarity: $s_v(q, f) = \cos(\mathbf{e}_q, \mathbf{e}_f)$.
  \item \textbf{Keyword ($R_k$):} SQLite FTS5 BM25 lexical matching with optional query expansion for synonyms and inflections.
  \item \textbf{Temporal ($R_t$):} Date-range queries over fact validity windows: $\{f \mid f.\texttt{valid\_from} \leq t_{\text{end}} \wedge (f.\texttt{valid\_until} = \bot \vee f.\texttt{valid\_until} \geq t_{\text{start}})\}$.
  \item \textbf{Entity/graph ($R_g$):} Entity-scoped retrieval traversing the knowledge graph from extracted query entities.
  \item \textbf{Support channels:} Bounded raw-episode rescue for exact or episodic questions and entity-scoped inference support for questions that require cautious reasoning from related facts.
\end{itemize}

The planner is a hand-coded rule table, not a learned classifier and not an LLM call. \Cref{tab:planner-rules} shows the main rules used in the reported LoCoMo runs; unmatched questions use the configured default channels, usually vector and keyword search.

\paragraph{Configuration boundary.} The reported traces use frozen configurations whose route weights, retrieval budgets, answer policies, and enabled rescue paths are recorded with the artifacts. This boundary is reproducibility bookkeeping: it states exactly what was run and prevents silent configuration drift across result tables. No gold answers, rubric nuggets, judge labels, or benchmark scores are provided to the retrieval path or answer model. The retrieval routes described here are generic memory mechanisms rather than answer templates: the writer stores typed facts, and the reader retrieves those typed facts under the same deterministic query planner for ordinary user data.

\begin{table}[H]
\centering
{\small
\begin{tabularx}{\linewidth}{@{}l l X@{}}
\toprule
\textbf{Query signal} & \textbf{Main weights} & \textbf{Additional behavior} \\
\midrule
Inference / causal & inference support 2.0; entity 2.0; vector 1.5 & Pull up to 120 facts for the asked person and return up to 8 deduplicated support facts. \\
Contradiction / update & keyword 2.0; temporal 2.0; vector 1.5; entity 1.5 & Include superseded facts so current and prior claims can be compared. \\
Multi-session / relation & graph 3.0; entity 2.0; vector 1.5 & Expand graph traversal depth for cross-session evidence. \\
Summary / recap & entity 2.5; keyword 2.0; graph 1.5 & Use all configured channels and a larger context budget. \\
Explicit date bound & temporal 3.0; keyword 2.0; entity 2.0 & Parse date ranges and filter by fact validity windows. \\
List / aggregation & entity 3.0; vector 1.5; keyword 1.5 & Enable raw-episode rescue for list, count, exact, interval, and recommendation questions. \\
\bottomrule
\end{tabularx}}
\caption{Deterministic query-planner rules. Values are channel weights used by weighted RRF; omitted channels use the base weight or are inactive for that query.}
\label{tab:planner-rules}
\end{table}

Each route $R_c$ returns a ranked list of candidate facts. Candidates are fused using \emph{weighted reciprocal rank fusion} (RRF)~\citep{rrf}:
\begin{equation}
  \text{RRF}(f) = \sum_{c \in \mathcal{C}} \frac{w_c}{k + \text{rank}_c(f)}
  \label{eq:rrf}
\end{equation}
where $\mathcal{C}$ is the set of active channels, $\text{rank}_c(f)$ is the rank of fact $f$ in channel $c$ (or $\infty$ if absent), and $k$ is a smoothing constant. The default configuration uses $k=20$.

\paragraph{Post-fusion filters.} After RRF scoring, two deterministic filters adjust rankings:

\emph{Person demotion:} If the query targets person $p$, facts whose \texttt{scope\_entity} $\neq p$ receive a demotion factor $\delta_p = 0.05$:
\begin{equation}
  s'(f) = \begin{cases}
    \text{RRF}(f) & \text{if } f.\texttt{scope\_entity} = p \text{ or unscoped} \\
    \text{RRF}(f) \cdot \delta_p & \text{otherwise}
  \end{cases}
  \label{eq:person-filter}
\end{equation}

\emph{Preservation floor:} If fact $f$ is already a high-ranked candidate and matches the asked person and temporal window, it cannot drop below rank $r_{\min} = 25$ after reranking, preventing validated candidates from being buried by noisy high-scoring candidates.

The preservation floor is a packing-protection mechanism, not a truth-resolution mechanism. It is applied after active-state filtering, so it assumes the write path has already marked outdated state facts as superseded. In this sense, preservation is \emph{lifecycle-trusting}: it protects active candidates from reranking loss, but it does not decide which of two active state facts is the current truth. Eywa's intended invariant is that each mutable state slot has at most one active fact for a given \texttt{user\_id}, \texttt{scope\_entity}, and \texttt{fact\_type}; stale values should be handled by write-time supersession rather than by read-time filtering.

\paragraph{Route preservation and rescue.} When a top-ranked candidate from a single route also matches the asked person, time window, and query terms, Eywa preserves a bounded number of those candidates even if later reranking would bury them. Exact, list, count, interval, and recommendation-shaped questions may also activate raw-episode rescue. This channel searches the original episode FTS index, reranks up to 24 candidate episodes by query-term overlap, speaker match, and lexical score, and returns up to 6 short excerpt facts capped at 8 source lines and roughly 1{,}300 characters. Inference support is separate: it activates only for inference, causal, or supported-reasoning queries with a resolved person, retrieves up to 120 facts for that person, embeds fact plus source text, scores candidates by $0.7$ cosine similarity and $0.3$ lexical overlap, and returns up to 8 deduplicated support facts.

\paragraph{Context packing.} The final ranked list is packed into a context string $\mathcal{C}_q$ under a configured token budget $B$. Facts are grouped by entity when possible and then packed in ranked order; diversity or temporal ordering modes are used for summary, ordering, and time-range queries. The packing function is:
\begin{equation}
  \mathcal{C}_q = \text{Pack}\!\left(\text{top-}K\!\left(\{f \mid s'(f) > 0\}\right),\; B\right)
  \label{eq:packing}
\end{equation}

\subsection{Answer Policy Separation and Traceability}

The \texttt{ReadResult} returned by the memory system contains two distinct fields:

\begin{itemize}[leftmargin=*,topsep=2pt,itemsep=1pt]
  \item \texttt{context}: The retrieved memory evidence, formatted for the answer model.
  \item \texttt{answer\_instructions}: Policy text that tells the answer model how to use the evidence. This is \emph{not} injected into the memory context.
\end{itemize}

Three answer modes are available:

\begin{description}[style=nextline,leftmargin=24mm,labelwidth=22mm,topsep=2pt,itemsep=1pt]
  \item[\texttt{strict}:] Answer only from direct evidence. Safest for small or low-capability models that tend to hallucinate when given license to infer.
  \item[\texttt{balanced}:] Allow clear implications from one strong memory or multiple consistent memories, without outside knowledge.
  \item[\texttt{advanced}:] Allow cautious reasoning and ordinary common-knowledge bridges when retrieved memories provide specific support. Designed for strong-reasoning models.
\end{description}

This separation is critical because the same retrieval architecture can produce different outcomes depending on the answer model and policy. A local 3B model in \texttt{advanced} mode may hallucinate; a frontier model in \texttt{strict} mode may refuse despite adequate evidence. The answer policy makes this a product decision rather than an architectural limitation.

Formally, the \texttt{ReadResult} can be summarized as:
\begin{equation}
  \texttt{ReadResult}(q) = \bigl(\mathcal{C}_q,\; \mathcal{I}_\pi,\; F_q,\; D_q,\; t_{\text{ms}}\bigr)
  \label{eq:readresult}
\end{equation}
where $\mathcal{C}_q$ is the packed context string, $\mathcal{I}_\pi$ is the answer instruction text for policy $\pi \in \{\text{strict},\allowbreak \text{balanced},\allowbreak \text{advanced}\}$, $F_q$ is the set of rendered fact identifiers, $D_q$ contains route counts, route errors, and query-plan diagnostics, and $t_{\text{ms}}$ is the retrieval latency.

The same structure makes failures diagnosable. If support is absent from the memory store, the issue is a coverage or extraction gap. If support exists but does not reach $\mathcal{C}_q$, the issue is retrieval, ranking, or packing. If support reaches $\mathcal{C}_q$ but the answer is wrong or refused, the issue is answer policy or model capability. If the answer is semantically supported but scored wrong, the issue is measurement. This trace connects the failure taxonomy in \cref{tab:failure-taxonomy} to concrete system artifacts rather than treating benchmark errors as a single undifferentiated score.

\section{Evaluation}

Eywa is evaluated as an inspectable memory system rather than as a single end-to-end prompt. For each benchmark question, the trace preserves the question, gold answer or rubric, retrieved context, answer instructions, model answer, judge label or score, and run configuration. This lets aggregate benchmark scores be read together with the artifacts that produced them, rather than as opaque final numbers. LoCoMo, LongMemEval-S, and BEAM are therefore used as complementary probes of different memory behaviors rather than averaged into one leaderboard number. LoCoMo tests long-horizon conversational recall and reasoning; LongMemEval-S tests retrieval sufficiency across memory categories; BEAM stresses technical memory, updates, ordering, abstention, and rubric-level partial credit.

\subsection{Benchmark Setup}

We report three completed benchmark families in this version: LoCoMo, LongMemEval-S, and BEAM. LoCoMo and LongMemEval-S are reported with their own judge-accuracy or retrieval-sufficiency protocols. BEAM uses a rubric-nugget protocol, so we report it separately as raw mean nugget score plus pass@score$\geq$0.5 rather than mixing it into the LoCoMo comparison table.

\paragraph{LoCoMo.}
LoCoMo~\citep{locomo} evaluates very long-term conversational memory over 10 multi-session conversations. The public release contains 1{,}986 QA questions across five categories: 841 single-hop, 282 multi-hop, 321 temporal reasoning, 96 open-domain knowledge, and 446 adversarial questions. For headline comparison, we follow the published result-table convention and report the non-adversarial C1--C4 split only: 1{,}540 answerable questions across single-hop, multi-hop, temporal, and open-domain categories. We report each LoCoMo run by three model roles:

\begin{itemize}[leftmargin=*,topsep=2pt,itemsep=1pt]
  \item \textbf{Write-path model:} the model used to extract or validate memories during ingestion.
  \item \textbf{QA model:} the model that receives Eywa's retrieved context and answer instructions.
  \item \textbf{Judge model:} the model that scores final answers with the benchmark judge prompt.
\end{itemize}

We report C1--C4 judge accuracy as the primary LoCoMo metric. We report official-style token F1 only where the corresponding trace export retained it. The adversarial C5 subset is useful for safety analysis, but it is not included in the headline LoCoMo comparison tables in this paper version.

We distinguish two result types. \emph{Write--QA aligned} runs use the same model family for writing memory and answering questions, with GPT-4o used as the judge. These are the primary LoCoMo results because they avoid mixing a strong extractor with a weak answerer, or vice versa. \emph{Cross-role diagnostic} runs intentionally mix write and QA roles to isolate a specific system question; for example, running Qwen3 32B as the QA model over a Sonnet-written memory store measures how much a stronger write path helps a smaller answer model. All LoCoMo rows in the main tables use frozen, artifact-recorded run configurations whose retrieved contexts and scoring labels are preserved with the public traces.

\paragraph{LongMemEval-S.}
LongMemEval-S~\citep{longmemeval} is a 500-question benchmark designed around five core abilities: information extraction, multi-session reasoning, knowledge updates, temporal reasoning, and abstention. The released question-type labels further distinguish single-session user, single-session assistant, single-session preference, multi-session, temporal-reasoning, and knowledge-update cases; abstention questions are marked by question id. The LongMemEval-S run reported here uses GPT-4o for the write path, QA, and judging over a 3{,}200-token retrieval budget. Because LongMemEval uses different tasks and a different binary judge protocol, we report it separately from the LoCoMo comparison tables.

\paragraph{BEAM.}
BEAM is a technical-memory stress suite used in this paper to test detailed recall, updates, ordering, abstention, and multi-session synthesis. It contains 35 conversations and 700 rubric-scored questions across 10 categories. Because BEAM is introduced here rather than inherited as an established public leaderboard, we treat it as a stress evaluation and report it separately from LoCoMo and LongMemEval-S. The accompanying research release is intended to include the conversations, questions, rubric nuggets, retrieved contexts, model answers, and judge outputs so that readers can inspect the benchmark at the question level. The run reported here uses Kimi K2.5 hybrid ingestion, Claude Sonnet 4.6 for QA, and Claude Sonnet 4.6 for rubric judging over Eywa's retrieved context. BEAM questions are scored by rubric nuggets rather than a single exact answer, so we report the raw mean nugget score, pass@score$\geq$0.5, pass@score$\geq$0.8, and perfect-score counts. The 0.5 threshold is a coarse sufficiency threshold rather than a claim of exact correctness; higher thresholds are included to expose partial-answer behavior.

\paragraph{Baseline numbers.} For context, we include published or artifact-reported C1--C4 scores from other systems evaluated on LoCoMo~\citep{truememory,hindsight,hindsightbench,memos,memoseval}. The closest external headline rows in this literature context are True Memory Pro and Hindsight Gemini-3, but their protocols differ from Eywa's trace-preserving runs: True Memory reports a semantic-judge mean without this table's category breakdown, while Hindsight uses a different model and judge protocol. The remaining external rows are taken from the MemOS evaluation artifact. These numbers are therefore provided as literature context rather than a controlled head-to-head reproduction; direct comparison is limited by differences in extraction models, answer models, judge prompts, judge model versions, retrieval budgets, and reported trace formats.

\begin{table}[H]
\centering
{\scriptsize
\begin{tabularx}{\linewidth}{@{}p{0.13\linewidth} p{0.16\linewidth} p{0.24\linewidth} p{0.15\linewidth} X@{}}
\toprule
\textbf{Benchmark} & \textbf{Questions / split} & \textbf{Model roles} & \textbf{Retrieval budget} & \textbf{Metric and artifact} \\
\midrule
LoCoMo & 1{,}540 C1--C4 questions from 10 conversations & Write and QA roles vary by run; aligned runs use the same family; judge is GPT-4o & Frozen artifact-recorded retrieval configuration; trace stores packed context & Judge accuracy; per-question artifacts include question, gold answer, model answer, retrieved context, and label. \\
LongMemEval-S & 500 questions across released question-type buckets & GPT-4o write, QA, and judge & 3{,}200 tokens & Retrieval-sufficiency judge accuracy; failure buckets retained for diagnostic analysis. \\
BEAM & 700 rubric-scored questions across 35 technical-memory conversations & Kimi K2.5 write; Claude Sonnet 4.6 QA; Claude Sonnet 4.6 rubric judge & Deep technical retrieval profile used for the canonical run & Mean nugget score plus pass@score$\geq$0.5; per-question rubric, answer, score, and retrieved context retained. \\
\bottomrule
\end{tabularx}}
\caption{Evaluation protocol summary. Reported metrics are not averaged across benchmarks because LoCoMo, LongMemEval-S, and BEAM use different scoring conventions.}
\label{tab:evaluation-protocol}
\end{table}

\paragraph{Reproducibility artifacts.}
The accompanying research bundle publishes per-question traces rather than only aggregate scores. For the reported runs, artifacts include the benchmark question, gold answer or rubric nuggets, retrieved context, model answer, scoring label or nugget score, and the result file used to compute the table. Public-facing artifacts are intended to be accessible from \url{https://eywa.to/research}, making Eywa's benchmark claims independently inspectable at the question level.

\subsection{LoCoMo Results}

\begin{table}[H]
\centering
{\scriptsize
\begin{tabularx}{\linewidth}{@{}>{\raggedright\arraybackslash}X l l l r r@{}}
\toprule
\textbf{Run} & \textbf{Write} & \textbf{QA} & \textbf{Judge} & \textbf{Correct / total} & \textbf{Accuracy} \\
\midrule
\textbf{Claude Sonnet 4.6} & Sonnet 4.6 & Sonnet 4.6 & GPT-4o & 1{,}389 / 1{,}540 & \textbf{90.19\%} \\
GPT-4o & GPT-4o & GPT-4o & GPT-4o & 1{,}367 / 1{,}540 & 88.77\% \\
Kimi K2.5 & Kimi K2.5 & Kimi K2.5 & GPT-4o & 1{,}295 / 1{,}540 & 84.09\% \\
\bottomrule
\end{tabularx}}
\caption{Measured Eywa LoCoMo C1--C4 write--QA aligned runs under frozen artifact-recorded configurations. All rows use GPT-4o as judge and are therefore the main LoCoMo comparison rows.}
\label{tab:main-results}
\end{table}

\Cref{tab:main-results} summarizes the current measured LoCoMo results using role-normalized reporting. These write--QA aligned runs are the main comparison because they keep judge protocol fixed while changing the model family used to write and answer over memory. The gap between rows shows that final benchmark quality depends on extraction quality, answer-model reasoning, and judging, not only on the retrieval substrate.

The full per-question Sonnet 4.6 LoCoMo result, including questions, gold answers, model answers, retrieved context, and pass/fail labels, is published at \url{https://eywa.to/research}.

\begin{table}[H]
\centering
{\scriptsize
\begin{tabularx}{\linewidth}{@{}X r r r@{}}
\toprule
\textbf{LoCoMo trace} & \textbf{Mean context tokens} & \textbf{Median} & \textbf{p95} \\
\midrule
Sonnet 4.6 write + Sonnet 4.6 QA & 4{,}257 & 4{,}338 & 5{,}975 \\
Qwen3 32B write + Qwen3 32B QA & 3{,}177 & 3{,}223 & 4{,}407 \\
\bottomrule
\end{tabularx}}
\caption{Retrieved-memory context footprint for two LoCoMo C1--C4 traces. Counts are artifact-recorded memory-context tokens supplied to the QA model and exclude question text, answer instructions, answer generation, and judge calls.}
\label{tab:locomo-context-tokens}
\end{table}

\begin{table}[H]
\centering
{\scriptsize
\begin{tabularx}{\linewidth}{@{}p{0.24\linewidth} p{0.28\linewidth} p{0.18\linewidth} X@{}}
\toprule
\textbf{Run} & \textbf{Roles} & \textbf{Result} & \textbf{Purpose} \\
\midrule
Qwen3 32B full cycle & Write: Qwen3 32B; QA: Qwen3 32B; scoring: dashboard reviewed state & 1{,}064 / 1{,}540 = 69.09\% & Same-family diagnostic baseline for the Qwen3 32B write and answer loop. The gap to the Sonnet-written memory row isolates write-path quality as a material factor. \\
Qwen3 32B on Sonnet memory & Write: Sonnet 4.6; QA: Qwen3 32B; scoring: reviewed QA pass & 1{,}227 / 1{,}540 = 79.68\% & Measures whether a smaller answer model benefits from a stronger write path. Not a leaderboard-comparable GPT-4o-judge row. \\
\bottomrule
\end{tabularx}}
\caption{LoCoMo Qwen3 32B diagnostic runs. These diagnostics isolate model-role effects and are reported separately from GPT-4o-judged write--QA aligned comparisons.}
\label{tab:cross-role-diagnostic}
\end{table}

The Qwen3 32B diagnostic shows that write-path quality and evidence organization materially affect smaller answer models, while the remaining C3 weakness reflects answer-model reasoning rather than simple memory absence.

\begin{table}[H]
\centering
{\scriptsize
\begin{tabular}{@{}lrrrr@{}}
\toprule
\textbf{Run} & \textbf{C1 Multi-hop} & \textbf{C2 Temporal} & \textbf{C3 Reasoning} & \textbf{C4 Single-hop} \\
\midrule
\textbf{Claude Sonnet 4.6} & \textbf{93.97\%} & \textbf{89.41\%} & 78.13\% & \textbf{90.61\%} \\
GPT-4o & 88.65\% & 88.47\% & \textbf{82.29\%} & 89.66\% \\
Kimi K2.5 & 87.59\% & 81.31\% & 77.08\% & 84.78\% \\
Qwen3 32B full cycle & 57.09\% & 68.85\% & 30.21\% & 77.65\% \\
Qwen3 32B on Sonnet memory & 84.04\% & 79.75\% & 41.67\% & 82.52\% \\
\bottomrule
\end{tabular}}
\caption{LoCoMo category accuracy by model configuration. The Qwen3 32B rows are diagnostic dashboard-reviewed runs and are not GPT-4o-judged comparison rows; their overall pass rates are reported in \cref{tab:cross-role-diagnostic}.}
\label{tab:locomo-category}
\end{table}

\Cref{tab:locomo-category} shows two effects. First, Eywa remains strong across answer-model regimes: even the weaker Qwen3 32B answer model remains competitive on single-hop and temporal recall when it reads a Sonnet-written memory store. Second, C3 open-domain reasoning is the most model-sensitive category. This is expected: many C3 questions require inference over retrieved evidence, so the memory layer can surface the relevant facts while a smaller answer model may still fail to connect them.

The Qwen3 32B diagnostics separate write-path quality from answer-model quality. When Qwen3 32B writes and answers over its own memory store, the reviewed C1--C4 pass rate is 1{,}064 / 1{,}540 = 69.09\%. When the same Qwen3 32B answer model reads a Sonnet-written memory store, the reviewed pass rate rises to 1{,}227 / 1{,}540 = 79.68\%. We therefore report these rows as diagnostics rather than controlled GPT-4o-judged leaderboard rows.

The Kimi and Qwen configurations are important for deployment: the same retrieval architecture remains useful outside a single frontier-model setup. This supports Eywa's product thesis that high-quality memory should not require an LLM call on every read, while still leaving room for stronger write or answer models when applications need multi-hop or open-domain reasoning.

\begin{table}[H]
\centering
{\scriptsize
\begin{tabular}{@{}lrrrrr@{}}
\toprule
\textbf{Method} & \textbf{Single-hop} & \textbf{Multi-hop} & \textbf{Temporal} & \textbf{Open-domain} & \textbf{Overall} \\
\midrule
True Memory Pro$^\dagger$ & -- & -- & -- & -- & 93.00\% \\
Hindsight Gemini-3$^\ddagger$ & 86.17\% & 70.83\% & 83.80\% & 95.12\% & 89.61\% \\
MemOS & 81.09\% & 67.49\% & 75.18\% & 55.90\% & 75.80\% \\
memobase & 73.12\% & 64.65\% & 81.20\% & 53.12\% & 72.01\% \\
Mem0 & 73.33\% & 58.75\% & 52.34\% & 45.83\% & 64.57\% \\
MIRIX & 68.22\% & 54.26\% & 68.54\% & 46.88\% & 64.33\% \\
Zep & 66.23\% & 52.12\% & 54.82\% & 33.33\% & 59.22\% \\
MemU & 66.34\% & 63.12\% & 27.10\% & 50.00\% & 56.55\% \\
Supermemory & 67.30\% & 51.12\% & 31.77\% & 42.67\% & 55.34\% \\
\bottomrule
\end{tabular}}
\caption{External LoCoMo C1--C4 literature context from recent memory-system result tables. Sources: True Memory Pro from \citet{truememory}; Hindsight from \citet{hindsight,hindsightbench}; remaining rows from the MemOS evaluation artifact~\citep{memos,memoseval}. Eywa's measured rows are reported separately in \cref{tab:main-results,tab:locomo-category}. $^\dagger$True Memory reports a 3-run semantic-judge mean and does not provide this table's category breakdown. $^\ddagger$Hindsight's category labels are mapped to the same four LoCoMo categories, but its model and judge protocol differ. External rows are not controlled reproductions and should not be read as a single leaderboard.}
\label{tab:published-context}
\end{table}

\subsection{LongMemEval-S Results}

\begin{table}[H]
\centering
{\small
\begin{tabular}{@{}lrr@{}}
\toprule
\textbf{Question type} & \textbf{Questions} & \textbf{Judge} \\
\midrule
Knowledge update & 78 & 96.2\% \\
Multi-session & 133 & 84.2\% \\
Single-session assistant & 56 & 73.2\% \\
Single-session preference & 30 & 90.0\% \\
Single-session user & 70 & 100.0\% \\
Temporal reasoning & 133 & 87.2\% \\
\midrule
\textbf{Overall} & \textbf{500} & \textbf{88.2\%} \\
\bottomrule
\end{tabular}}
\caption{LongMemEval-S full 500-question GPT-4o retrieval-sufficiency result by question type.}
\label{tab:longmemeval-results}
\end{table}

\Cref{tab:longmemeval-results} shows a second benchmark setting with GPT-4o used consistently across write-path, QA, and judge roles. Eywa is strongest on direct user-memory questions and knowledge updates, while assistant-memory, multi-session, and temporal-reasoning questions remain harder.

\begin{table}[H]
\centering
{\small
\begin{tabular}{@{}lr@{}}
\toprule
\textbf{Failure bucket} & \textbf{Count} \\
\midrule
Fact not retrieved & 27 \\
Candidate lost before final context & 12 \\
Fact not extracted & 12 \\
Assembly required across multiple memories & 7 \\
Packed evidence not used by answer model & 1 \\
\midrule
\textbf{Total remaining failures} & \textbf{59} \\
\bottomrule
\end{tabular}}
\caption{LongMemEval-S remaining failures after GPT-4o judging.}
\label{tab:longmemeval-failures}
\end{table}

\Cref{tab:longmemeval-failures} shows that the remaining LongMemEval errors are no longer dominated by simple single-fact lookup. The largest residual bucket is retrieval coverage, followed by context-selection loss and extraction gaps. This gives Eywa a concrete improvement path for the next iteration.

\subsection{BEAM Results}

BEAM is reported as a stress evaluation rather than a LoCoMo-style comparison. Each question is judged against rubric nuggets, so a partial answer can receive partial credit. \Cref{tab:beam-results} reports the completed 35-conversation, 700-question canonical run, using Kimi K2.5 hybrid ingestion and Claude Sonnet 4.6 for both QA and rubric judging.

\begin{table}[H]
\centering
{\scriptsize
\begin{tabular}{@{}lrrrrr@{}}
\toprule
\textbf{BEAM category} & \textbf{N} & \textbf{Mean score} & \textbf{Pass@0.5} & \textbf{Pass@0.8} & \textbf{Perfect} \\
\midrule
Abstention & 70 & 92.86\% & 65 / 70 & 65 / 70 & 65 / 70 \\
Contradiction resolution & 70 & 93.21\% & 70 / 70 & 52 / 70 & 52 / 70 \\
Event ordering & 70 & 79.49\% & 60 / 70 & 50 / 70 & 39 / 70 \\
Information extraction & 70 & 81.43\% & 58 / 70 & 55 / 70 & 53 / 70 \\
Instruction following & 70 & 76.67\% & 58 / 70 & 48 / 70 & 48 / 70 \\
Knowledge update & 70 & 70.00\% & 49 / 70 & 49 / 70 & 49 / 70 \\
Multi-session reasoning & 70 & 87.77\% & 63 / 70 & 58 / 70 & 57 / 70 \\
Preference following & 70 & 79.05\% & 60 / 70 & 48 / 70 & 48 / 70 \\
Summarization & 70 & 64.07\% & 49 / 70 & 38 / 70 & 28 / 70 \\
Temporal reasoning & 70 & 90.00\% & 65 / 70 & 61 / 70 & 61 / 70 \\
\midrule
\textbf{Overall} & \textbf{700} & \textbf{81.45\%} & \textbf{597 / 700} & \textbf{524 / 700} & \textbf{500 / 700} \\
\bottomrule
\end{tabular}}
\caption{BEAM technical-memory stress result by rubric category under the deep technical retrieval profile; configuration is recorded in the artifacts. Mean score is the average rubric-nugget score; pass@0.5, pass@0.8, and perfect-score counts are shown separately because BEAM allows partial credit.}
\label{tab:beam-results}
\end{table}

The BEAM result is useful because it stresses detailed technical recall, update handling, ordering, abstention, and multi-session synthesis in a format that exposes partial failures. Eywa is strongest on contradiction resolution, abstention, temporal reasoning, multi-session reasoning, and information extraction. Summarization and knowledge update remain the clearest next memory-improvement targets, but the rubric view shows that knowledge update is no longer a collapse point.

\paragraph{Operational metrics.} Eywa's default memory read path uses no LLM calls before the answer model receives context. On a stress memory store with 6{,}320 facts, 250 episodes, and 14{,}585 fact-entity links, the default interactive retrieval profile returns assembled context in roughly 150--200 ms. The system can optionally perform deeper provenance assembly for audit and benchmark diagnostics, but the product-facing path is the fast LLM-free retrieval profile reported in \Cref{tab:retrieval-profiles}.

\begin{table}[H]
\centering
{\small
\begin{tabularx}{\linewidth}{@{}l X r@{}}
\toprule
\textbf{Retrieval mode} & \textbf{What it does} & \textbf{Observed latency} \\
\midrule
Hot vector search & In-memory vector index lookup before full context assembly. & 1--3 ms internal \\
Vector returned context & Vector-only retrieval with assembled returned context. & 148--171 ms \\
Interactive retrieval & Vector plus keyword retrieval with entity, temporal, and lifecycle filters. & 198--200 ms \\
Startup / initialization & Loads stores, graph checkpoint, and hot vector index. & 823--885 ms \\
\bottomrule
\end{tabularx}}
\caption{Default LLM-free retrieval latency on the stress memory store. Latencies exclude answer generation and judging.}
\label{tab:retrieval-profiles}
\end{table}

For debugging and audit workflows, Eywa can also assemble deeper provenance context from raw evidence. That diagnostic path is intentionally separate from the default interactive retrieval profile and motivates future work on source-text indexing and evidence-chunk storage.

Zep/Graphiti also preserves provenance through temporal graph structure. Eywa's provenance claim is narrower: raw evidence is stored separately from derived beliefs, and two-tier canonical writes maintain explicit evidence-to-fact links so a retrieved fact can be traced back to the source turn and supporting signals. Observation-mode facts additionally retain source text and episode metadata.

\paragraph{Closest-system contrast.} Eywa is closest in spirit to systems that combine extraction, retrieval, and temporal structure, but it draws the system boundary differently. Compared with Zep/Graphiti, Eywa emphasizes immutable source evidence as a first-class object separate from graph-derived beliefs and answer instructions. Compared with Mem0-style drop-in memory, Eywa treats extraction as an index over preserved evidence rather than as the memory itself. Compared with A-MEM-style note enrichment, Eywa spends model work on the write path but keeps retrieval deterministic and LLM-free. Compared with leaderboard-style memory reports, Eywa's evaluation claim is tied to per-question traces that expose the retrieved context and judge outcome behind each aggregate score.

\begin{table}[H]
\centering
{\scriptsize
\setlength{\tabcolsep}{3pt}
\begin{tabularx}{\linewidth}{@{}p{0.14\linewidth}p{0.10\linewidth}p{0.17\linewidth}p{0.13\linewidth}p{0.12\linewidth}p{0.15\linewidth}p{0.10\linewidth}@{}}
\toprule
\textbf{System} & \textbf{Read LLM} & \textbf{Write LLM} & \textbf{Provenance} & \textbf{Temporal} & \textbf{Retrieval} & \textbf{Policy split} \\
\midrule
Mem0 & n.s. & multi-stage & n.s. & n.s. & hybrid & n.s. \\
Mem0$^g$ & n.s. & multi-stage + graph & n.s. & partial & graph + semantic & n.s. \\
Supermemory$^\dagger$ & n.s. & n.s. & partial & partial & API/RAG & n.s. \\
Zep/Graphiti & n.s. & LLM extraction & graph-linked & bi-temporal & graph search & n.s. \\
A-MEM & n.s. & multi-step & note-linked & n.s. & cosine & n.s. \\
MemoryOS & n.s. & LLM update/summarize & n.s. & n.s. & tiered & n.s. \\
MemGPT & tool-mediated & tool-mediated & n.s. & n.s. & tool-call & n.s. \\
\midrule
\textbf{Eywa} & \textbf{0} & \textbf{1} & evidence-linked & temporal metadata & 4-route & separated \\
\bottomrule
\end{tabularx}}
{\footnotesize n.s. = not publicly specified in the cited paper or documentation reviewed here. Sources: Mem0 and Mem0$^g$ from \citet{mem0paper}; Supermemory from public product documentation~\citep{supermemory}; Zep/Graphiti from \citet{zep}; A-MEM from \citet{amem}; MemoryOS from \citet{memoryos}; MemGPT from \citet{memgpt}. External write/read fields are qualitative public-source summaries rather than measured call counts, and pipeline stages are not directly comparable across papers and product docs. Provenance semantics differ across systems.}
\caption{Architectural context for agent memory systems. The table summarizes public claims and papers; it is not a controlled implementation audit. Eywa's row reports the implementation used in this paper's artifacts.}
\label{tab:arch-comparison}
\end{table}

\subsection{Refusal-Aware Evaluation}
\label{sec:refusal-aware}

In our evaluation harness, the official-style LoCoMo F1 scorer can award credit when the model refuses and the gold answer text is not found verbatim in the retrieved context. This behavior is useful for adversarial questions, but it can inflate scores for answerable categories when a model abstains instead of using paraphrased or indirect evidence.

We define \emph{refusal-aware F1}. Let $\hat{y}$ be the system prediction, $y^*$ the gold answer, and $\text{cat}(q)$ the question category:
\begin{equation}
  \text{RA-F1}(q) = \begin{cases}
    \text{F1}(\hat{y}, y^*) & \text{if } \neg\text{refusal}(\hat{y}) \\
    0 & \text{otherwise}
  \end{cases}
  \label{eq:refusal-aware}
\end{equation}
This metric measures whether the system actually answers, not whether it avoids answering. For answerable categories, refusal-aware F1 assigns zero to non-adversarial refusals.

We keep this metric in the trace format because it is useful for engineering analysis, especially on open-domain inference questions where abstention can hide answer-model weakness. It is not used as the primary headline score in this version; the reported LoCoMo and LongMemEval-S results use role-explicit judge accuracy or retrieval-sufficiency accuracy, BEAM uses rubric-nugget scoring, and token-level F1 is reported separately where available.

\subsection{Model Separation}
\label{sec:answer-ablation}

The role-normalized runs in \Cref{tab:main-results,tab:cross-role-diagnostic} demonstrate the value of model separation. The same retrieval architecture serves frontier, budget, and smaller answer-model regimes, while the trace format keeps write-path quality and answer-model behavior distinguishable. The Qwen diagnostics are especially useful because they show the difference between a weak write path and a weak answerer over a stronger memory store.

This validates the design choice: rather than building the answer model's capability into the memory system, Eywa returns the same evidence and lets operators choose the right model for their constraints. Privacy-sensitive deployments can run entirely local. Cost-sensitive deployments can use a budget API. Performance-critical deployments can use a frontier model. The memory layer does not need to change.

\subsection{Diagnostic Analysis}
\label{sec:failure-analysis}

\begin{figure}[t]
\centering
\includegraphics[width=0.95\linewidth]{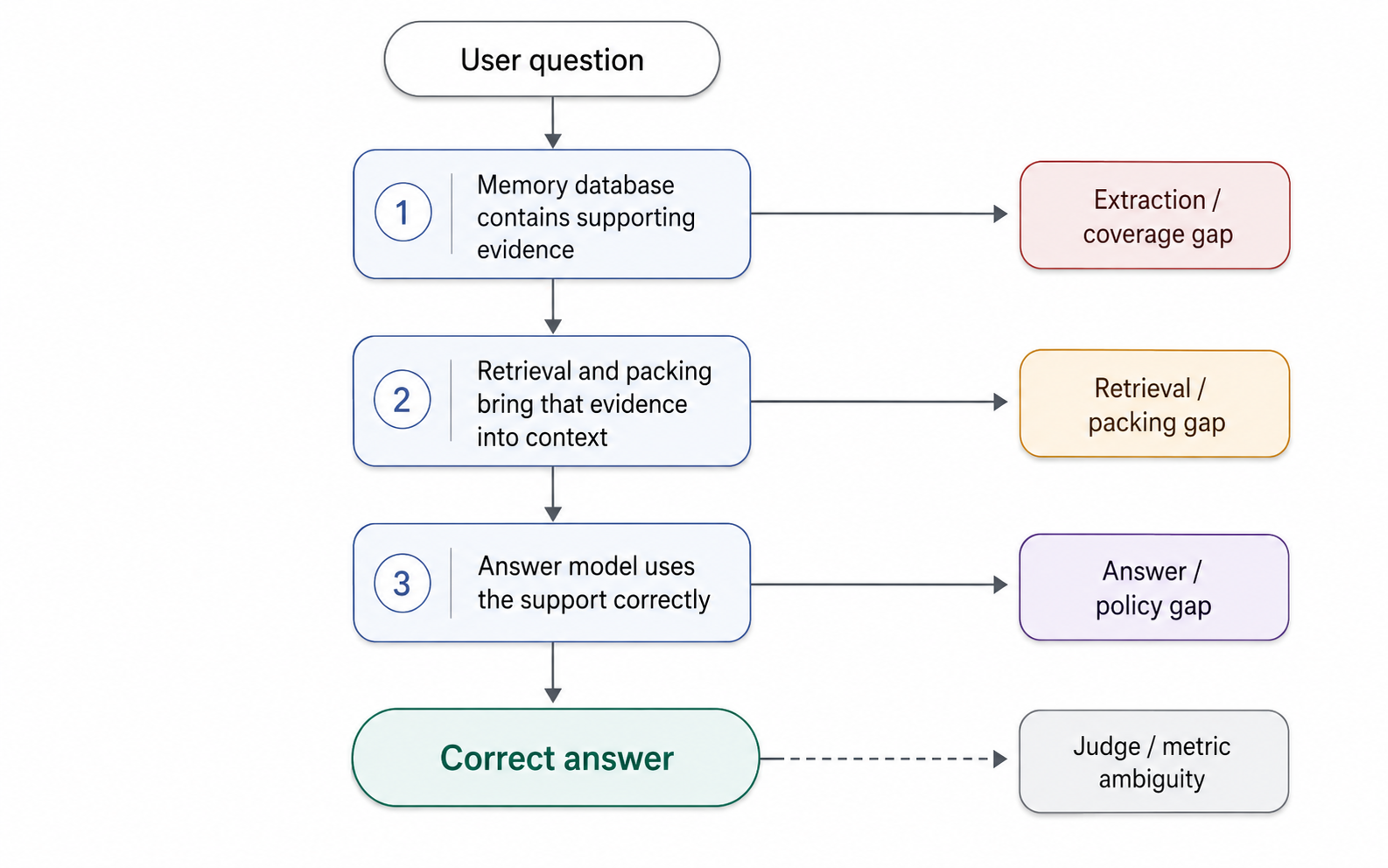}
\caption{Diagnostic funnel used for trace audits. The funnel is a method for localizing errors, not an additional benchmark score.}
\label{fig:funnel}
\end{figure}

Eywa records enough trace information to localize failures after a run: which facts were retrieved by each route, which survived fusion and packing, what answer instructions were provided, and whether the judge accepted the final answer. This makes benchmark errors inspectable rather than forcing all failures into a single score.

The current diagnostics show two signals. First, the same memory substrate remains usable across several model choices, which supports separating memory delivery from answer synthesis. Second, C3 open-domain questions remain the most difficult answerable LoCoMo category because many require inference over retrieved evidence rather than direct fact recall. We therefore treat C3 as the main target for future profile synthesis and inference-support improvements.

\section{Discussion}

\paragraph{Memory as infrastructure.}
The central implication of Eywa is that long-term memory should be treated as infrastructure rather than as prompt decoration. A production agent may change answer models, deployment targets, privacy requirements, or product policies over time, but its memory substrate should remain inspectable and reusable. Eywa's separation of evidence, beliefs, retrieval, and answer policy gives this substrate a stable interface: the write path preserves and validates what was said, the read path returns bounded evidence, and the answer model verbalizes that evidence under a chosen policy.

\paragraph{Why provenance matters for adoption.}
Users do not experience memory failures as benchmark errors. They experience them as broken trust: the agent remembers the wrong person, forgets a recent event, repeats stale information, or cannot explain where a claim came from. Provenance turns these failures into inspectable system states. If a fact is wrong, the source turn can be checked. If a user requests erasure, evidence can be deleted by user scope. If a benchmark answer fails, the trace can distinguish extraction, retrieval, packing, answer-policy, and judging errors. This is the difference between a memory feature and a memory system.

\paragraph{Local-first does not mean weak memory.}
The Kimi K2.5 and Qwen3 32B runs are important because they show that useful memory is not restricted to one frontier-only deployment pattern. This does not eliminate the value of stronger write or answer models. It does show that the memory layer can carry a large part of the system behavior, allowing users to choose between local privacy, low operating cost, and frontier reasoning without replacing the memory store.

\paragraph{Zero-LLM reads change the cost model.}
Many agent systems place LLM calls on both write and read paths. Eywa makes a different tradeoff: spend intelligence when memory is written, then serve memory through deterministic retrieval at read time. This is attractive for products because reads are often far more frequent than writes. A user may ask many questions over the same memory store, and each query should not require an additional memory-planning model call before the answer model is invoked. Deterministic reads also make latency, caching, and debugging easier to reason about.

\paragraph{Open source as a trust requirement.}
Memory is a sensitive layer: it stores personal preferences, work history, relationships, plans, and sometimes private facts. For this reason, the core memory system should be inspectable, portable, and usable without a hosted dependency. Eywa's open-source direction follows from the architecture itself. The strongest version of the system is not a locked feature tier; it is a shared substrate that developers can run locally, audit, extend, and connect to whichever answer model their deployment requires. Managed services can add scale, collaboration, analytics, and enterprise controls, but the core memory capability should remain genuinely useful on its own.

\paragraph{Refusal-aware evaluation in practice.}
We do not argue that refusals are always wrong. On C5 (adversarial), refusals \emph{are} the correct answer. We argue that non-adversarial refusals should not receive credit in aggregate metrics, because they indicate the system failed to answer a question it should have been able to answer. Separately tracking refusal rates alongside accuracy gives operators a clearer picture of system behavior, especially when the same memory evidence is consumed by models with different reasoning and safety profiles.

\section{Threats to Validity}

The benchmark results in this paper should be interpreted through their protocols rather than as a single universal score. LoCoMo C1--C4 accuracy, LongMemEval-S retrieval-sufficiency accuracy, and BEAM rubric-nugget scoring measure different behaviors and are not averaged. Published competitor rows also differ in extraction models, answer models, judge prompts, judge model versions, retrieval budgets, and whether they report strict matching, semantic judging, or rubric scoring. We therefore use external numbers as context, not as a controlled head-to-head reproduction.

Model roles are another source of variation. Eywa reports write-path, QA, and judge models explicitly because a stronger extractor, a stronger answer model, or a more permissive judge can all move final accuracy. The write--QA aligned LoCoMo rows are the main comparison rows because they share GPT-4o judging. Qwen3 32B diagnostics are reported separately and should not be read as equivalent to the GPT-4o-judged comparison rows.

The reported aggregate percentages also omit formal uncertainty estimates in this version. LoCoMo has only 10 conversations, so question-level percentages can overstate certainty if many errors are correlated within a conversation. Future versions should include confidence intervals, conversation-level bootstrap, and paired tests such as McNemar or paired bootstrap where two systems answer the same questions.

Finally, C5 adversarial LoCoMo questions are excluded from the headline C1--C4 comparison split, following the comparison convention used by recent result tables, but they remain important for safety and false-premise behavior. BEAM uses partial-credit rubric nuggets, so its mean score and pass thresholds expose different failure modes than binary judged accuracy. Because BEAM is introduced here, it should be read as a public stress-test contribution once released, not as an independently established community benchmark. These protocol boundaries are why the paper publishes per-question artifacts rather than asking readers to trust aggregate numbers alone.

\section{Limitations}

\begin{enumerate}[leftmargin=*,topsep=3pt,itemsep=2pt]
  \item \textbf{Extraction model dependency:} Our main results use LoCoMo memory stores created with high-recall extraction configurations. We report several model regimes, including Qwen-based runs, but we have not performed an exhaustive sweep over local extractors, answer models, retrieval budgets, and judge prompts.
  \item \textbf{Benchmark coverage:} We report the standard LoCoMo C1--C4 comparison split, a full LongMemEval-S run, and a full BEAM run. BEAM uses nugget scoring rather than simple judge accuracy, so it is reported as a separate stress evaluation instead of being averaged into the LoCoMo comparison table.
  \item \textbf{Benchmark configuration:} No gold answers, rubric nuggets, judge labels, or benchmark scores are provided to Eywa at retrieval or answer time. Reported runs use frozen configurations whose route weights, retrieval budgets, answer policies, and enabled rescue paths are preserved with the trace artifacts.
  \item \textbf{BEAM validation:} BEAM is introduced in this paper as a technical-memory stress suite. The release is intended to include conversations, questions, rubric nuggets, retrieved contexts, model answers, and judge outputs, but this version does not yet report independent human validation of BEAM rubrics, inter-annotator agreement, or external baseline reproductions.
  \item \textbf{Model roles:} We report write-path, QA, and judge models explicitly. Write--QA aligned runs reduce role ambiguity, but they are still not controlled head-to-head comparisons against external systems.
  \item \textbf{Judge validity:} The main LoCoMo rows use GPT-4o judging, and the canonical BEAM run uses Claude Sonnet 4.6 as both QA model and rubric judge. We publish traces and prompts for inspection, but we have not yet reported multi-judge agreement, blinded human audit, or judge-disagreement analysis.
  \item \textbf{Ablations and significance:} This version reports completed system runs and diagnostic failures, but not a full ablation suite over canonical-only, observation-only, hybrid extraction, raw rescue, entity routing, temporal routing, graph traversal, preservation floors, or answer-policy modes. It also does not include confidence intervals or statistical significance tests for the LoCoMo gaps.
  \item \textbf{Planner and hyperparameters:} Query-shape rules, channel weights, $k=20$, $\delta_p=0.05$, and $r_{\min}=25$ are fixed hand-coded configuration values selected from diagnostic runs, not learned parameters. We report route and post-fusion diagnostics, but not a full sensitivity sweep over these constants.
  \item \textbf{Intent routing:} The evaluated read path uses deterministic query-shape rules. Exemplar-based multilingual intent routing is ongoing shadow-mode work and is not claimed as part of the reported benchmark configuration.
  \item \textbf{State supersession invariant:} The preservation floor relies on the write path to keep mutable state clean. If two conflicting state facts remain active because a fact key was missing or contradiction detection did not fire, preservation may protect the lexically stronger stale candidate. Production deployments should enforce and periodically audit the invariant that each \texttt{user\_id}, \texttt{scope\_entity}, and mutable \texttt{fact\_type} has at most one active state fact.
  \item \textbf{Projection consistency:} The local-first implementation treats SQLite as authoritative and LanceDB plus graph state as idempotent projections. This avoids claiming an atomic distributed transaction across all stores. Projection rebuild and startup reconciliation are deliberate parts of the evaluated single-process design; hosted multi-process deployment is a separate systems extension rather than a claim made by the current implementation.
  \item \textbf{Signal detector coverage:} Hard anchors are sparse in conversational memories. In the 143-sample two-tier audit, 67.4\% of candidate facts had no hard anchor; in the LoCoMo observation database used for the main run, only 59 of 2{,}541 persisted facts contained a hard anchor in the fact text. Eywa therefore treats hard-anchor checks as one guardrail inside a broader support validator, not as proof that every zero-anchor fact is semantically verified. We have not yet measured token-level recall for every detector type.
  \item \textbf{Adversarial safety:} The adversarial LoCoMo C5 subset is not included in the headline comparison tables in this version. Safety and false-premise handling remain partly policy-dependent and should be reported separately in deployment evaluations.
  \item \textbf{Revision evaluation:} Eywa represents corrections, rejections, approvals, and superseding facts in the write path, but LoCoMo does not isolate revision handling as a dedicated category. A targeted update-and-supersession benchmark is needed to evaluate this capability directly.
  \item \textbf{Latency:} Reported retrieval latencies describe the default interactive read path and exclude answer and judge generation. Diagnostic provenance assembly over raw evidence is available for audit workflows, but it is not the default product retrieval profile. Latency may differ under production hardware, cache state, and database size.
  \item \textbf{Scale and erasure:} The evaluated system is local-first and single-process. We describe evidence deletion and projection rebuild as architectural requirements, but do not yet report end-to-end erasure tests across projections, logs, traces, backups, or hosted multi-tenant deployments. Scale results beyond the reported stress store are future systems work.
\end{enumerate}

\section{Future Work}

Eywa's evaluated deployment profile is local-first and single-process: SQLite is the authoritative store, while the vector index and graph state are rebuildable idempotent projections. Extending this design to horizontally scaled hosted infrastructure requires explicit distributed-systems work: an outbox or projection journal, background reconciliation, stronger cross-process locking, and operational monitoring for projection drift. This is future work, not a hidden assumption behind the reported benchmark results.

Additional future work includes source-evidence FTS, evidence-chunk storage for audit workflows, sensitivity sweeps for route weights and preservation floors, a factorial model-role matrix, multi-judge and human-audit subsets, C5/refusal-rate tables, ablations for retrieval routes and answer policies, relational-query evaluation for multi-person questions, documented deployment-profile runs, and targeted benchmarks for mutable-state supersession.

\section{Conclusion}

Eywa demonstrates that separating evidence capture from canonical fact writing, and memory delivery from answer policy, produces a memory system that is both accurate and auditable. Across LoCoMo, LongMemEval-S, and BEAM, the reported runs show that the same memory substrate can serve different answer-model regimes while preserving traceability to the evidence behind each answer. The diagnostic traces show where errors arise across extraction, retrieval, packing, answer policy, and judging, making further improvement a measurable engineering process rather than guesswork.

The refusal-aware analysis exposes a systematic inflation risk in official-style LoCoMo scoring, particularly on inference-heavy C3 questions where abstention can hide answer-model weakness. We encourage future memory system papers to report refusal rates and refusal-aware metrics alongside accuracy.

Eywa is designed to run on local hardware and to be released as an inspectable memory substrate. Its direction is to become a practical default memory layer for AI agents: auditable enough for research, useful enough for local-first deployment, and structured enough to grow into production and hosted settings with additional systems validation. Project and benchmark artifacts are intended to be published at \url{https://eywa.to}.

\bibliographystyle{plainnat}
\bibliography{eywa}

\appendix
\setlength{\parindent}{0pt}
\setlength{\parskip}{6pt plus 1pt minus 1pt}

\section{Trace Fields}
\label{app:trace-fields}

\begin{table}[H]
\centering
{\small
\begin{tabularx}{\linewidth}{@{}lX@{}}
\toprule
\textbf{Field} & \textbf{Purpose} \\
\midrule
Question metadata & Conversation id, category, question text, gold answer, and evidence identifiers when provided by the benchmark. \\
Retrieved candidates & Per-route candidate facts, route ranks, fused scores, entity scope, temporal metadata, and source channels. \\
Packed context & Final context delivered to the answer model, with fact identifiers preserved for provenance inspection. \\
Answer policy & The selected policy mode and generated answer instructions, stored separately from memory context. \\
Model output & Raw answer text, refusal flag, F1 score, refusal-aware F1 score, judge label, and judge rationale when available. \\
Latency & Read-path timing separate from answer and judge generation. \\
\bottomrule
\end{tabularx}}
\caption{Trace fields saved for benchmark auditing.}
\label{tab:trace-fields}
\end{table}

\section{Implementation Details}
\label{app:implementation}

\begin{table}[H]
\centering
\begin{tabularx}{\linewidth}{@{}lX@{}}
\toprule
\textbf{Component} & \textbf{Implementation} \\
\midrule
Canonical store & SQLite (WAL mode), single-writer lock \\
Vector index & LanceDB + hot in-memory index \\
Graph store & RustworkX (serialized, not thread-safe) \\
Keyword search & BM25 with query expansion \\
Embeddings & ONNX Runtime (local, no API call) \\
Extraction LLM & Configurable (GPT-4o-mini, Qwen, and other supported models) \\
Signal detection & Rule-based (regex + spaCy NER) \\
Query planner & Deterministic (no LLM call) \\
\bottomrule
\end{tabularx}
\caption{Eywa technology stack.}
\label{tab:stack}
\end{table}

\end{document}